\documentclass[10pt,twocolumn,letterpaper]{article}

\usepackage[pagenumbers]{cvpr} 

\usepackage{multirow}
\usepackage[accsupp]{axessibility}  

\definecolor{cvprblue}{rgb}{0.21,0.49,0.74}
\usepackage[pagebackref,breaklinks,colorlinks,allcolors=cvprblue]{hyperref}

\def\paperID{HiGen-12} 
\def\confName{CVPR}
\def\confYear{2026}

\title{Improving Human Image Animation via Semantic Representation Alignment}

\author{
Chang Liu$^{1}$, Mengting Chen$^{2}$, Yixuan Huang, Haoning Wu$^{1}$,\\
Chen Ju$^{2}$, Shuai Xiao$^{2}$, Jinsong Lan$^{2}$, Yanfeng Wang$^{1}$\\
$^{1}$School of Artificial Intelligence, Shanghai Jiao Tong University, China\\
$^{2}$Alibaba Group, China
}

\begin{document}
\maketitle
\begin{abstract}
The field of image-to-video generation has made remarkable progress. 
However, challenges such as human limb twisting and facial distortion persist, especially when generating long videos or modeling intensive motions.
Existing human image animation works address these issues by incorporating human-specific semantic representations,~({\em e.g.}, dense poses or ID embeddings) as additional conditions.
However, conditioning on these representations could decrease the generation flexibility.
Moreover, their reliance on RGB pixel supervision also lacks emphasis on learning necessary 3D geometric relationships and temporal coherence.
In contrast, we introduce a novel approach named \textbf{SemanticREPA} that leverages these semantic representations as supervision signals through representation alignment.
Specifically, we begin by training a \textbf{structure alignment} module that aligns the structure representations obtained from video latents with video depth estimation features.
We then fix the pretrained module, and utilize it to provide additional supervision on the structure representations of the diffusion models, achieving structure rectification to generate coherent and stable human structures.
Simultaneously, we develop an \textbf{ID alignment} module to align the ID representations of the generated videos to face recognition features. We further propose to use the predicted structure representations to refine identity restoration in relevant regions. 
With structure and ID alignment, our method demonstrates superior quality on extended character motions and enhanced character consistency.
\end{abstract}    
\begin{figure*}
  \includegraphics[width=\textwidth]{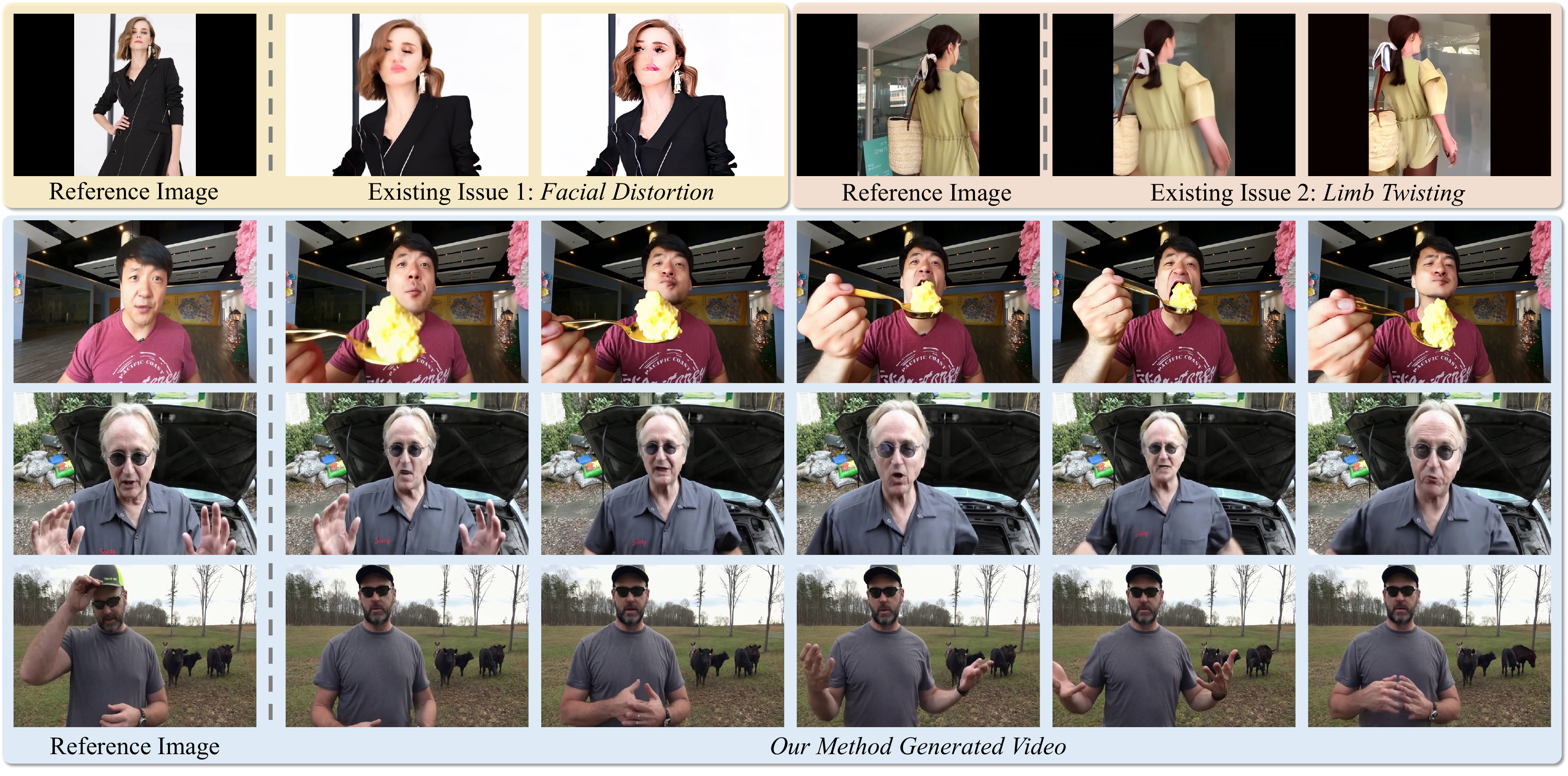}
  \caption{In human image animation task, most existing image-to-video models exhibit issues like \textit{human limb twisting} and \textit{facial distortion} in generated videos. Our SemanticREPA employs \textit{semantic representation alignment} as additional supervision during fine-tuning, allowing the generation of character motion videos with stable human structures and consistent identities.
  }
  \label{fig:teaser}
\end{figure*}

\section{Introduction}

In recent years, diffusion models~\cite{ho2020denoising,song2020denoising} have emerged as the leading approach in image generation.
Models such as Stable Diffusion~\cite{SDM,SD3}, DALL-E~\cite{dalle1,dalle2,dalle3}, and Imagen~\cite{imagen1,imagen2,imagen3} have achieved remarkable results. With the advent of scalable diffusion transformer architectures~\cite{Peebles2022DiT} and large-scale datasets~\cite{nan2024openvid,chen2024panda70m}, diffusion models have extended their success from image generation to the realm of video generation,
for example, MovieGen~\cite{moviegen}, CogVideoX~\cite{yang2024cogvideox} and Open-Sora~\cite{opensora}, along with proprietary commercial models like OpenAI Sora~\cite{sora} and Kling~\cite{kling}. Utilizing the latest models, users can now rapidly generate high-quality videos with hundreds of frames guided by various conditioning inputs.

Human image animation, as a specialized adaptation of image-to-video generation, involves generating videos of a single person's motion from an image featuring only one individual, guided by text or other conditions.
Although most recent works emphasize incorporating additional motion controls in the animation process, such as pose guidance~\cite{hu2023animateanyone,xu2023magicanimate,2024mimicmotion,zhu2024champ}, flow guidance~\cite{chen2023mcdiff,niu2024mofa,shi2024motion,ma2024cinemo}, 
and camera trajectory guidance~\cite{li2024imageconductor,wang2024humanvid}, 
they still rely on RGB pixel-level supervision and suffer from issues such as limb twisting and facial distortion.
This severely undermines human structure stability and character consistency, particularly when generating long videos with extensive movements.
Consequently, improving the backbone of image-to-video generative models for human image animation remains a challenging, yet highly valuable problem.

In this paper, we consider to leverage semantic representation alignment as additional supervision, thereby enhancing stability and consistency in image-to-video generation. To mitigate limb distortion, we propose aligning the structure representations of video latents to video depth estimation features~\cite{chen2025video}, while for facial distortion, we consider aligning the ID representations to face recognition features~\cite{deng2019arcface,paraperas2024arc2face}.
At training time, our method first trains the alignment modules for structure rectification and ID restoration, respectively, predicting the corresponding semantic representations from compressed VAE~\cite{kingma2013auto} latents.
Next, we freeze the alignment modules and further fine-tune the diffusion transformer backbone. Consequently, the image-to-video backbone achieves strong understanding for 3D human motion and temporal identity consistency, enabling the generation of higher-quality human motion videos. As shown in Figure~\ref{fig:teaser}, our method could generate long videos with stable human structures and consistent identities, avoiding facial distortion and limb twisting.

To summarize, we make the following contributions in this paper:
(i) We propose a solution SemanticREPA to limb twisting and facial distortion issues in diffusion-based human image animation models by leveraging semantic representation alignment;
(ii) We develop two alignment modules that predict corresponding semantic representations directly from video latents through knowledge distillation, which enables to leverage semantic representations as supervisions instead of conditions;
(iii) Quantitative and qualitative evaluations confirm the effectiveness of our method, demonstrating its ability to generate long character motion videos with improved structure stability and character consistency.
\section{Related Works}

\subsection{Diffusion-based Video Generation}
With the tremendous progress of diffusion models in image generation~\cite{SDM, SD3}, recent research has increasingly focused on developing diffusion-based video generation models.
Early works~\cite{guo2023animatediff,blattmann2023videoldm,wang2023modelscope,girdhar2024emuvideo} incorporate decoupled spatial and temporal layers based on pretrained Stable Diffusion text-to-image models.
Stable Video Diffusion~\cite{blattmann2023stable} advances further by introducing the first open-source image-to-video model, capable of generating videos up to 25 frames long.
Some studies~\cite{2023i2vgenxl,wang2023lavie,bartal2024lumiere} also explore using cascaded latent diffusion models to enhance performance.
With the introduction of the scalable DiT~\cite{Peebles2022DiT} architecture, recent works~\cite{hong2022cogvideo,ma2024latte,bao2024vidu,opensora} has increasingly favored transformer-based architectures, as their parameter scaling better accommodates the growing size of video datasets~\cite{nan2024openvid,chen2024panda70m}.
Models such as CogVideoX~\cite{yang2024cogvideox}, MovieGen~\cite{moviegen}, and Kling~\cite{kling} can generate tens to hundreds of video frames in a single run, guided by text or image prompts.
Our work aims to optimize the image-to-video backbone in the human portrait domain, addressing issues such as facial distortion and limb artifacts.

\subsection{Diffusion-based Human Image Animation}
Human image animation requires models to generate videos of a single given character's movements, guided by image prompts and other conditions like text, denoting an adaptation of the image-to-video task in the human portrait domain. 
Given its substantial application potential and commercial value, this task has attracted considerable research attention.
Recent literature has primarily focused on incorporating additional motion conditioning techniques into the image-to-video generation process.
LivePhoto~\cite{chen2023livephoto} estimates motion intensity from text prompts.
Some studies~\cite{hu2023animateanyone,xu2023magicanimate,2024mimicmotion,zhu2024champ} employ sparse, dense or 3D pose sequences to constrain character movements in the video.
Other works~\cite{chen2023mcdiff,niu2024mofa,shi2024motion,ma2024cinemo} utilize or first predict optical flow to guide motion within the video, particularly in cases involving camera trajectory guidance~\cite{li2024imageconductor,wang2024humanvid}.
While these studies have introduced various motion conditioning techniques, their dependence on semantic representations as conditions restricts generative flexibility.
Moreover, their reliance on RGB pixel-level supervision hampers the learning of 3D geometric structures and physical consistency.
In contrast, our approach leverages semantic representations as supervision signals rather than conditions, aiming to enhance 3D geometric structure stability and temporal identity consistency without compromising flexibility.

\subsection{Diffusion-based Semantic Representation}
As a form of generalized self-supervised learning, the internal features of diffusion models can serve as highly effective semantic representations.
Research on diffusion models and semantic representations has primarily focused on two directions:
on one hand, some studies~\cite{ddae2023,chen2024deconstruct} leverage internal features of diffusion models to perform dense discriminative tasks, such as object segmentation~\cite{li2023guiding,xu2023odise} and monocular depth estimation~\cite{ke2023repurposing,he2024lotus}.
On the other hand, other studies~\cite{zhao2024diffree,yu2024reprealign,RCG2023,pernias2024wrstchen,jeong2024track4gen} propose to optimize the semantic representations within diffusion models, enhancing both training efficiency and generation quality.
REPA~\cite{yu2024reprealign} attempts to optimize image diffusion models by utilizing self-supervised semantic representations~\cite{oquab2023dinov2} through knowledge distillation.
Our work aligns more closely with the latter.
While recent studies remain on image generation models, we aim to extend these insights to the video generation domain, particularly for the task of human image-to-video animation.
A concurrent study~\cite{yang2025unified} simultaneously generates and supervises video generation models along with their corresponding segmentation masks or depth maps.
However, it neglects temporal identity consistency and requires modifications to the transformer backbone.
\section{Motivation}

To address the issues in human image animation, 
we consider to enhance the internal features of the diffusion transformer through semantic representation alignment,
aiming to achieve higher generation quality and consistency in long video generation eventually. In this section, we outline the motivations behind our methodology and training strategy.

We begin by presenting insights on \textit{how semantic representation alignment can mitigate the identified issues}.
Existing methods~\cite{hu2023animateanyone,xu2023magicanimate,2024mimicmotion,zhu2024champ}, while incorporating extra conditions, still primarily rely on RGB pixel-level supervision. 
Without explicit proxy tasks, they lack emphasis on learning 3D geometry, physical plausibility, or long-term consistency, making it difficult to maintain accurate spatial relationships and temporal coherence across extended sequences.
Our approach addresses these issues by introducing explicit proxy tasks including depth and identity supervision, to enforce the model to learn geometric structure and temporal consistency.
This targeted supervision helps the model encode meaningful spatial and temporal cues, resulting in better spatial fidelity and long-term consistency in generated videos.

Specifically, for \textbf{limb twisting}, artifacts such as distorted, blurred, or even disappearing limbs are commonly observed during movement, especially in fine-grained body parts like fingers or during rapid motion.
These artifacts stem from the model's limited capacity to accurately model 3D body movements.
To address this, we apply \textbf{structure representation alignment} to distill prior knowledge of 3D human motion into the diffusion transformer.
Structure representation alignment directs the supervision to focus primarily on the 3D human geometric structure, effectively mitigating the influence of texture information.
For \textbf{facial distortion}, the diversity and subtlety of human facial expressions often cause facial features in human image animation to shift with motion, gradually deviating from the original reference image.
We believe this distortion arises from the model's difficulty in maintaining fine-grained temporal consistency during extensive movement.
To mitigate this, we introduce \textbf{ID representation alignment} to explicitly supervise temporal identity consistency in generated videos.

A straightforward implementation might be to decode the video latents predicted by the diffusion transformer using a VAE and directly supervise the resulting RGB frames.
However, for long video generation, this method is impractical because the VAE decoding step incurs significant memory overhead to store gradients, even though the VAE parameters remain fixed during backpropagation. 
Therefore, we adopt a \textbf{two-stage training strategy} instead of directly supervising the RGB frames. In the first stage, we pretrain an alignment module using knowledge distillation to extract semantic representations from the VAE video latents. In the second stage, we fix the module and employ it to supervise the diffusion transformer.

\section{Method}
In this section, we start by formulating the problem of diffusion-based human image animation in Section~\ref{sec:problem}. 
Next, we describe the pretraining pipeline for the proposed alignment modules in Section~\ref{sec:pretraining}.
Finally, we provide details on fine-tuning the diffusion transformer with our semantic representation alignment supervision in Section~\ref{sec:training}.

\begin{figure*}[t]
  \centering
  \includegraphics[width=\textwidth]{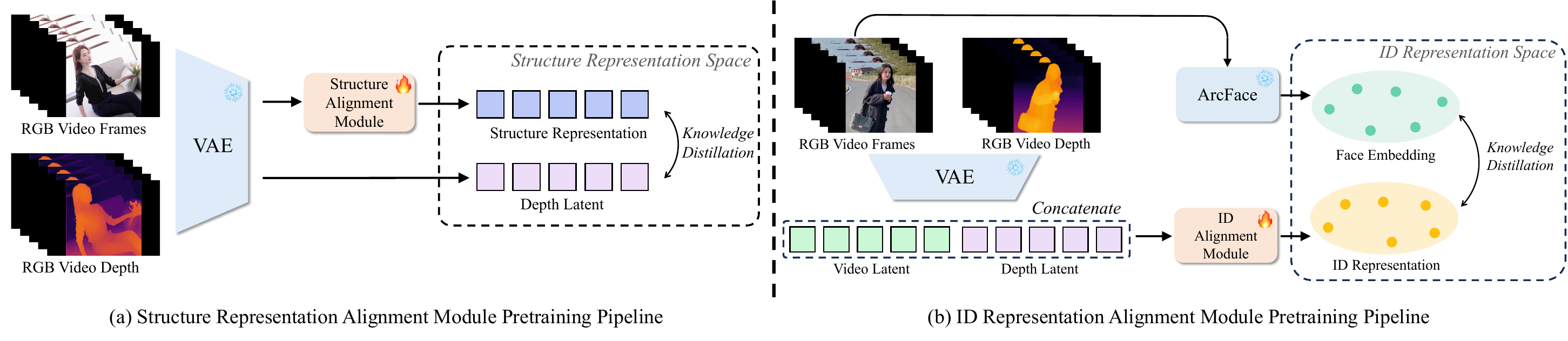} \\
  \caption{
  \textbf{Alignment Module Pretraining Pipeline Overview}. 
  (a) The Structure Alignment Module takes clean video latents as input, and outputs depth latents that align with the VAE-encoded RGB video depth.
  (b) The ID Alignment Module predicts facial representations based on video latents concatenated with depth latents, and aligns them with ArcFace features.
    }
 \label{fig:pretrain}
\end{figure*}

\subsection{Problem Definition}
\label{sec:problem}

In human image animation task, 
the generation process will be conditioned on a given reference image $I_{\text{ref}} \in \mathbb{R}^{H \times W \times 3} $ of the corresponding individual, and a text prompt $T$ describing the content of the human motion video. 
The objective is to generate a sequence of $N$ video frames, 
represented as $ \mathcal{V} = \{I_1, I_2, \dots, I_N\} $, 
where each frame $I_i \in \mathbb{R}^{H \times W \times 3}$. 
Here, $H$ and $W$ denote the height and width of each frame, respectively, and the generated video sequence should adhere to the appearance and context described by $I_{\text{ref}}$ and $T$.
The generation process is formulated as follows: 
\begin{align}
\label{eqn:hia}
\mathcal{V} = \{I_1, \ I_2, \ \dots, \ I_N\} = \Phi(I_{\text{ref}}, \ T; \ \Theta)
\end{align}
where $\Phi$ represents the diffusion-based human image animation model with trainable parameters $\Theta$.

\subsection{Alignment Module Pretraining}
\label{sec:pretraining}

In the following section, we provide a detailed description of the pretraining pipeline for the two alignment modules.

\subsubsection{Structure Alignment Module Pretraining}
\label{sec:SAMpretraining}
In addition to standard latent diffusion model training, we introduce an auxiliary task that predicts the structure representations from video latents.
Specifically, we formulate the structure representation prediction as a human-centric video depth estimation task.
For the structure alignment module, we use the CogVideoX transformer architecture~\cite{yang2024cogvideox} with a reduced number of layers, leveraging the fact that depth estimation and video generation are both dense prediction tasks.
The core of our alignment involves using RGB video latents to predict depth latents with the same noise level, which leads to two formulations: one that uses clean RGB latents as input and another that employs noisy ones.

As shown in Figure~\ref{fig:pretrain}, we begin by pretraining the structure alignment module with clean RGB latents as input. 
The module takes the noiseless video latent $\mathbf{z}_0$ as input and outputs the corresponding depth latent $\tilde{\mathbf{d}}_{0}(\mathbf{z}_{0})$.
\begin{align}
\label{eqn:sam}
\tilde{\mathbf{d}}_{0}(\mathbf{z}_{0}) = f_{\text{SAM}}(\mathbf{z}_0)
\end{align}
where \( f_{\text{SAM}} \) represents the structure alignment module.
We use Video Depth Anything~\cite{chen2025video} as the teacher model to extract pseudo ground truth video depth $\mathbf{D}$.
With brighter colors denoting closer distances, 
we colorize the single-channel video depth maps into RGB depth maps, and apply the VAE to encode the RGB depth $\mathbf{D}$ into the depth latent $\mathbf{d}_0$.
All other models remain fixed during the structure alignment module pretraining.
The training objective minimizes the MSE loss between the pseudo ground truth depth latent $\mathbf{d}_0$ and the predicted depth latent:
\begin{align}
\label{eqn:samloss}
\mathcal{L}_{\text{MSE}} = \left\| \mathbf{d}_0 - \tilde{\mathbf{d}}_{0}(\mathbf{z}_{0}) \right\|^2
\end{align}

In case of employing a noisy RGB latent $\mathbf{z}_t$ as input, our structure alignment module instead predicts the corresponding noisy depth latent $\tilde{\mathbf{d}}_{t}(\mathbf{z}_{t})$.
This prediction is conditioned on timestep $t$, and equations \ref{eqn:sam} and \ref{eqn:samloss} are transformed into:
\begin{align}
\label{eqn:sam1}
\tilde{\mathbf{d}}_{t}(\mathbf{z}_{t}) = f_{\text{SAM}}(\mathbf{z}_t, t)
\end{align}
\begin{align}
\label{eqn:samloss1}
\mathcal{L}_{\text{MSE}} = \left\| \mathbf{d}_t - \tilde{\mathbf{d}}_{t}(\mathbf{z}_{t}) \right\|^2
\end{align}

\begin{figure*}[t]
  \centering
  \includegraphics[width=\textwidth]{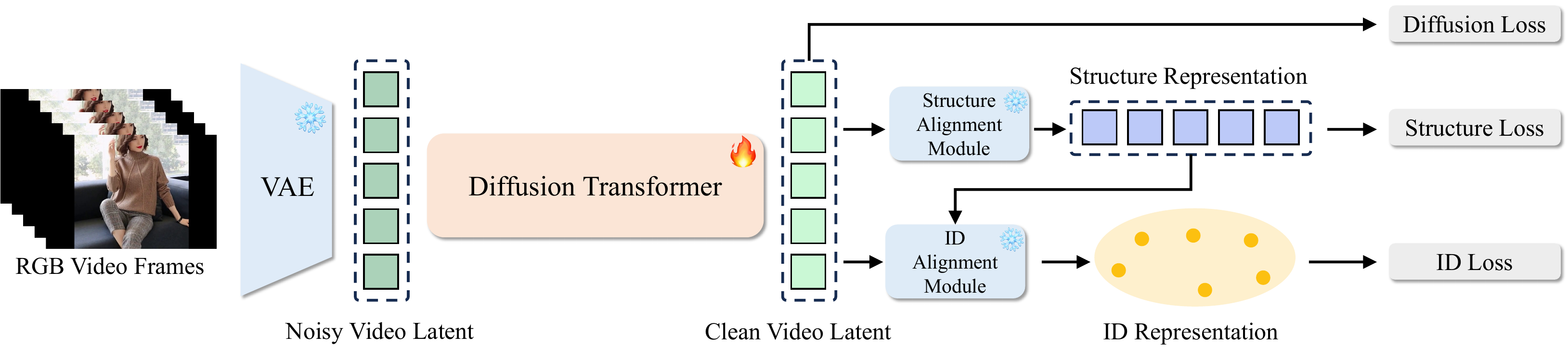} \\
  \caption{
  \textbf{Diffusion Transformer Fine-tuning Pipeline Overview.} With the assistance of the pretrained structure alignment module and ID alignment module, we apply additional supervision to the diffusion transformer fine-tuning through semantic representation alignment. We fix the two pretrained alignment modules, and only fine-tune diffusion transformer backbone.
    }
 \label{fig:finetune}
\end{figure*}

\subsubsection{ID Alignment Module Pretraining}
Simultaneously, we introduce another auxiliary task to predict ID representations from clean video latents, framing ID representation prediction as a face recognition feature extraction task.
Given the relative simplicity of this task, we employ a convolutional network composed of ResNet blocks~\cite{he2016deep} to extract ID representations. Since noisy video latents cannot effectively capture face recognition features, we use noiseless video latents as input for ID representation prediction.

As shown in Figure~\ref{fig:pretrain}, the input to our alignment module is the original video latent $\mathbf{z}_0$ concatenated with the ground truth depth latent $\mathbf{d}_0$, allowing it to predict the corresponding ID representations $\tilde{\mathbf{f}}(\mathbf{z}_0, \mathbf{d}_0)$ for each frame:
\begin{align}
\label{eqn:id}
\tilde{\mathbf{f}}(\mathbf{z}_0, \mathbf{d}_0) = f_{\text{ID}}(\mathbf{z}_0, \mathbf{d}_0)
\end{align}
where \( f_{\text{ID}} \) represents the ID alignment module.
During pretraining, we use the ground truth video latent $\mathbf{z}_0$ and depth latent $\mathbf{d}_0$ as input, while in fine-tuning, we will switch to the predicted results.
Since most videos only contain a single individual, we apply face detection to each video frame using the ArcFace model~\cite{deng2019arcface} to automatically locate the face and extract its features.
The detected facial features serve as the facial representation $\mathbf{f}$ for the reference image. 
The training objective is to minimize the L1 loss between the normalized ground truth face embedding $\mathbf{f}$ and the predicted face embedding $\tilde{\mathbf{f}}(\mathbf{z}_0, \mathbf{d}_0)$:
\begin{align}
\label{eqn:idloss}
\mathcal{L}_{1} = \left\| \mathbf{f} - \tilde{\mathbf{f}}(\mathbf{z}_0, \mathbf{d}_0) \right\|
\end{align}

\subsection{Diffusion Transformer Fine-tuning}
\label{sec:training}

After obtaining the two pretrained alignment modules, we will leverage them for applying additional supervision to the diffusion transformer fine-tuning through semantic representation alignment.
For a noisy video latent $\mathbf{z}_t \in \mathbb{R}^{B \times l \times c \times h \times w}$, the diffusion transformer predicts the added noise $\tilde{\boldsymbol{\epsilon}}_\theta(\mathbf{z}_t, t, \mathbf{c})$ based on the text condition $\mathbf{c}$, and calculates the corresponding original video latent $\tilde{\mathbf{z}}_0$. 
The scheduler then denoises for one step and obtains $\tilde{\mathbf{z}}_{t-1}$.
In case of clean RGB depth input, the structure alignment module takes $\tilde{\mathbf{z}}_0$ as input to predict the clean depth latent $\tilde{\mathbf{d}}_{0}(\tilde{\mathbf{z}}_0) \in \mathbb{R}^{B \times l \times c \times h \times w}$ corresponding to the input.
Simultaneously, the colorized RGB depth $\mathbf{D}$ will be encoded by VAE to obtain the ground truth latents $\mathbf{d}_{0}$.
The structure loss $\mathcal{L}_{S}$ can be expressed as the MSE loss between $\mathbf{d}_{0}$ and $\tilde{\mathbf{d}}_{0}(\tilde{\mathbf{z}}_0))$:
\begin{align}
\label{eqn:depthloss}
\mathcal{L}_{\text{S}} = \|\mathbf{d}_{0} - \tilde{\mathbf{d}}_{0}(\tilde{\mathbf{z}}_0)\|^2 
\end{align}

If we employ the noisy depth formulation, we exploit $\tilde{\mathbf{z}}_{t-1}$ as input and the structure loss $\mathcal{L}_{S}$ will be expressed as:
\begin{align}
\label{eqn:depthloss1}
\mathcal{L}_{\text{S}} = \|\mathbf{d}_{t-1} - \tilde{\mathbf{d}}_{t-1}(\tilde{\mathbf{z}}_{t-1}, t)\|^2 
\end{align}

The ID representation alignment module takes the predicted clean video latent $\tilde{\mathbf{z}}_0$ concatenated with the predicted depth latent $\tilde{\mathbf{d}}_{0}(\tilde{\mathbf{z}}_0)$ as input and predicts the ID representation $\tilde{\mathbf{f}}(\tilde{\mathbf{z}}_0, \tilde{\mathbf{d}_0}) \in \mathbb{R}^{B \times l \times 512}$. 
The ground truth ID representation $\mathbf{f}$ is computed by the alignment module using $\mathbf{z}_0$ and $\mathbf{d}_0$ as input. The ID loss $\mathcal{L}_{ID}$ is formulated as the L1 loss between $\mathbf{f}$ and $\tilde{\mathbf{f}}(\tilde{\mathbf{z}}_0, \tilde{\mathbf{d}_0})$:
\begin{align}
\label{eqn:faceloss}
\mathcal{L}_{\text{ID}} = \|\mathbf{f} - \tilde{\mathbf{f}}(\tilde{\mathbf{z}}_0, \tilde{\mathbf{d}_0})\|
\end{align}

For the noisy RGB depth formulation in the structure alignment module, we directly use the ground truth $\mathbf{d}_{0}$ to replace $\tilde{\mathbf{d}}_{0}(\tilde{\mathbf{z}}_0)$ to calculate $\mathcal{L}_{ID}$.
We omit the alternative expression of $\mathcal{L}_{ID}$ for simplicity.

In summary, the final loss $\mathcal{L}_{\text{final}}$ is a weighted sum of the diffusion loss $\mathcal{L}_t$, structure loss $\mathcal{L}_{\text{S}}$, and ID loss $\mathcal{L}_{\text{ID}}$, which can be expressed as follows:
\begin{align}
\label{eqn:loss}
\mathcal{L}_t = \mathbb{E}_{t \sim [1, T], \mathbf{z}_0, \boldsymbol{\epsilon}} \Big[\|\boldsymbol{\epsilon} - \boldsymbol{\epsilon}_\theta(\mathbf{z}_t, t, \mathbf{c})\|^2 \Big]
\end{align}
\begin{align}
\label{eqn:finalloss}
\mathcal{L}_{\text{final}} = \mathcal{L}_t + \lambda_{\text{S}} \mathcal{L}_{\text{S}} + \lambda_{\text{ID}} \mathcal{L}_{\text{ID}}
\end{align}

where $\lambda_{\text{S}}$, and $\lambda_{\text{ID}}$ are the weights of each respective loss component.
\section{Experiments}

In this section, we begin by outlining our experimental setup, including the dataset and implementation details in Section~\ref{sec:setup}.
We then compare our method with other models leveraging qualitative visualization and quantitative metrics in Section~\ref{sec:comparison}. 
Finally, we present ablation study results in Section ~\ref{sec:ablation} to demonstrate the effectiveness of our proposed semantic representation alignment supervision.

\begin{table*}[ht]
    \centering
    \begin{tabular}{l|ccccc|ccc}
        \toprule
        \multirow{2}{*}{Model} & \multicolumn{5}{c|}{Image-level Metrics} & \multicolumn{3}{c}{Video-level Metrics} \\
        & SSIM$\uparrow$ & PSNR$\uparrow$ & LPIPS$\downarrow$ & FID$\downarrow$ & CPBD$\downarrow$ & Motion Score$\downarrow$ & Text Score$\uparrow$ & ID Score$\uparrow$ \\
        \midrule     
        GT & - & - & - & - & 0.5347 & - & 0.2897 & 0.6465 \\
        \midrule
        VideoComposer & 0.1542 & 28.21 & 0.6721 & 1375.52 & 0.6444 & 55.01 & 0.2685 & 0.0423 \\
        I2VGen-XL & 0.1943 & 28.66 & 0.7467 & 1394.52 & 0.2075* & 67.10 & 0.2540 & 0.1492 \\
        DynamiCrafter & 0.3143 & 27.43 & 0.4889 & 2104.16 & 0.8333 & 62.81 & 0.2368 & 0.0246 \\
        SEINE & 0.3424 & 29.14 & 0.5275 & 556.29 & 0.6218 & 28.08 & 0.2879 & 0.1702\\            
        ConsistI2V & 0.7361 & 31.32 & 0.2811 & 924.55 & 0.4182* & 14.83 & 0.2728 & 0.0459 \\
        SVD & 0.3888 & 29.60 & 0.4590 & 467.95 & 0.3024* & 18.58 & 0.2778 & 0.3818 \\
        \midrule
        CogVideoX & \underline{0.7482} & \underline{32.40} & \textbf{0.1972} & \underline{247.37} & \underline{0.5839} & \underline{0.9426} & \underline{0.2942} & \underline{0.5087} \\
        SemanticREPA & \textbf{0.7502} & \textbf{32.51} & \underline{0.2011}  & \textbf{213.09} & \textbf{0.5817} & \textbf{0.4012} & \textbf{0.2956} & \textbf{0.6339} \\
        \bottomrule
    \end{tabular}
    \caption{\textbf{Quantitative Comparison with Other Baselines}. Our SemanticREPA outperforms all baselines across all metrics on our curated test set. \textbf{Bold} indicates the best results, and \underline{underlining} denotes the second-best. Values marked with * indicate less reliable CPBD scores.}
    \label{tab:model_comparison}
\end{table*}

\subsection{Experimental Setup}
\label{sec:setup}

\subsubsection{Dataset}
We fine-tune our model on the OpenVid-1M~\cite{nan2024openvid} dataset.
To filter for videos containing humans, we use YOLOv8~\cite{varghese2024yolov8} to perform human detection on the first frame of each video, retaining only those with detected humans.
Since most videos in OpenVid are TV-show-style with limited character movement, we supplement the dataset with in-house data consisting primarily of vertical-format try-on videos featuring more intensive character motion.
The OpenVid videos provide approximately 300K video-text pairs, while the in-house data contributes around 430K video-text pairs.
Given CogVideoX’s fixed resolution of 480×720 pixels, we sample 49 frames per video at 8 fps, resizing them to 480×720 resolution.
For vertical videos, we resize the height to 480 pixels and pad both sides to reach a width of 720 pixels.

\subsubsection{Implementation Details}
Our base model is CogVideoX 1.0~\cite{yang2024cogvideox}, which uses T5~\cite{raffel2020t5} as the text encoder.
For video depth feature extraction, we employ Video Depth Anything~\cite{chen2025video} for temporal consistent depth estimation.
For ID embedding extraction, we use the Arc2Face~\cite{paraperas2024arc2face} version of the ArcFace model~\cite{deng2019arcface} to obtain face recognition embeddings.
For further details, please refer to the supplementary material.

\subsection{Comparison with Other Methods}
\label{sec:comparison}


\subsubsection{Evaluation Metrics}
To evaluate the effectiveness of our proposed semantic representation alignment supervision, we collected a test set of 200 previously unseen videos with significant character motion for a fair comparison of generation quality and consistency across models.
We perform quantitative evaluation from both image-level quality and overall video-level quality perspectives.
For image-level metrics, following existing work, we use Structural Similarity Index (SSIM)~\cite{wang2004SSIM}, Peak Signal-to-Noise Ratio (PSNR)~\cite{hore2010PSNR}, Learned Perceptual Image Patch Similarity (LPIPS)~\cite{zhang2018LPIPS}, and Fr\'echet Inception Distance (FID)~\cite{heusel2017FID}. 
Additionally, we employ Cumulative Probability Blur Detection (CPBD)~\cite{narvekar2011cpbd} to assess the blur level in generated video frames.
For video-level metrics, we use the average ArcFace embedding cosine similarity as the ID score to measure ID consistency within the generated videos.
To compare motion modeling capabilities, we use RAFT~\cite{teed2020raft} to extract optical flow from the videos, colorize the flow maps, and calculate FID on these maps as the motion score, following PhysGen~\cite{liu2024physgen}.
We also utilize average CLIP~\cite{CLIP} cosine similarity to assess the alignment between generated videos and the text descriptions.
We do not use Fr\'echet Video Distance (FVD)~\cite{unterthiner2019fvd} due to the number of videos in our test set.

\subsubsection{Baseline Comparison}
We compare our SemanticREPA against current state-of-the-art image-to-video models, including VideoComposer~\cite{wang2024videocomposer}, I2VGen-XL~\cite{2023i2vgenxl}, DynamiCrafter~\cite{xing2024dynamicrafter}, SEINE~\cite{chen2023seine}, ConsistI2V~\cite{ren2024consisti2v}, and SVD~\cite{blattmann2023stable}.
Each of these models can generate videos based on a given first-frame image, with text prompts as optional conditions.
Since these UNet-based models cannot directly generate videos at the desired length and resolution, we apply a sliding window approach to obtain the full sequence of 49 frames and resize the generated frames to 480×720 pixels.
Additionally, we compare our model to the base CogVideoX without fine-tuning.

\subsubsection{Quantitative Evaluation Results}
The quantitative evaluation results are shown in Table~\ref{tab:model_comparison}.
Our SemanticREPA significantly outperforms all others across all evaluation metrics, achieving state-of-the-art performance.
Image-level metric results demonstrate that our model better captures the distribution of human motion images.
The CPBD score indicates that our generated images contain less blur, reflecting more stable human structures.
Notably, some models exhibit CPBD scores far below the ground truth, due to their inability to generate meaningful images over long video sequences, making their CPBD scores less reliable.
For video-level metrics, the motion score shows that our generated videos have motion patterns most similar to the ground truth distribution, while the ID score further verifies that our generated videos achieve the highest level of ID consistency.
These results demonstrate that our model significantly improves the human structure stability and ID consistency in generated videos without compromising other capabilities.

\subsubsection{Qualitative Evaluation Results}
We present the qualitative comparison results in the Supplementary Material, showing that our SemanticREPA generates videos with significantly better human structure stability and ID consistency, while other baselines fail to achieve this.
This strongly demonstrates our SemanticREPA's superior ability to model long character motions with enhanced consistency, effectively avoiding issues of facial distortion and limb twisting.

\begin{table*}[ht]
    \centering
    \begin{tabular}{lcc|ccccc|ccc}
        \toprule
        \multirow{2}{*}{Model} &\multirow{2}{*}{$\lambda_{\text{S}}$} &\multirow{2}{*}{$\lambda_{\text{ID}}$} & \multicolumn{5}{c|}{Image-level Metrics} & \multicolumn{3}{c}{Video-level Metrics} \\
        & & & SSIM$\uparrow$ & PSNR$\uparrow$ & LPIPS$\downarrow$ & FID$\downarrow$ & CPBD$\downarrow$ & Motion Score$\downarrow$ & Text Score$\uparrow$ & ID Score$\uparrow$ \\
        \midrule     
        GT & - & - & - & - & - & - & 0.5347 & - & 0.2897 & 0.6465 \\
        \midrule
        CogVideoX & - & - & 0.7482 & 32.40 & \underline{0.1972} & 247.37 & 0.5839 & 0.9426 & 0.2942 & 0.5087 \\
        CogVideoX\_F & - & - & \underline{0.7512} & 32.50 & \textbf{0.1968} & \underline{210.99} & 0.5839 & 0.7481 & \textbf{0.2959} & 0.5098 \\
        $\mathcal{L}_{\text{S}}(\mathbf{z}_t)$ & 0.1 & - & \textbf{0.7521} & \textbf{32.52} & 0.2009 & 214.22 & 0.5942 & 0.5123 & 0.2951 & 0.5987 \\
        $\mathcal{L}_{\text{S}}(\mathbf{z}_t)$ & 0.01 & - & \underline{0.7512} & \underline{32.51} & 0.2019 & 212.97 & 0.5922 & \underline{0.4427} & 0.2952 & 0.5751 \\
        $\mathcal{L}_{\text{S}}(\mathbf{z}_0)$ & 0.1 & - & 0.7489 & 32.47 & 0.2026 & \textbf{209.60} & 0.5911 & 0.7299 & 0.2950 & 0.6063 \\
        $\mathcal{L}_{\text{S}}(\mathbf{z}_0)$ & 0.01 & - & 0.7485 & \textbf{32.52} & 0.2026 & 212.21 & \underline{0.5837} & 0.5352 & 0.2953 & 0.6118 \\       
        $\mathcal{L}_{\text{ID}}$ & - & 10 & 0.7467 & 32.44 & 0.2056 & 226.54 & 0.5920 & 0.6338 & 0.2955 & \textbf{0.6451} \\
        $\mathcal{L}_{\text{ID}}$ & - & 1 & 0.7477 & 32.49 & 0.2045 & 219.90 & 0.5853 & 0.5397 & 0.2955 & 0.6223 \\
        $\mathcal{L}_{\text{ID}}$ \& $\mathcal{L}_{\text{S}}(\mathbf{z}_0)$ & 0.01 & 1  & 0.7502 & \underline{32.51} & 0.2011 & 213.09 & \textbf{0.5817} & \textbf{0.4012} & \underline{0.2956} & \underline{0.6339} \\
        \bottomrule
    \end{tabular}
    \caption{\textbf{Ablation Analysis on Supervision Implementations.} CogVideoX\_F stands for CogVideoX fine-tuned on our dataset.}
    \label{tab:ablation}
\end{table*}

\subsection{Ablation studies}
\label{sec:ablation}

\subsubsection{Structure Alignment Module Inputs}
As mentioned in Section~\ref{sec:SAMpretraining}, our structure alignment module has two implementations.
The first approach takes the clean RGB video latent as input.
Specifically, it utilizes the ground truth video latent $ \mathbf{z}_0 $ during alignment module pretraining and the transformer-predicted $ \tilde{\mathbf{z}}_0 $ during diffusion transformer fine-tuning. 
We denote this method as $ \mathcal{L}_{\text{S}}(\mathbf{z}_0) $.
The second approach employs the noisy video latents, with the $ t $-step noisy video latent $ \mathbf{z}_t $ during alignment module pretraining and the scheduler step result $ \tilde{\mathbf{z}}_{t-1} $ during fine-tuning.
We denote this method as $ \mathcal{L}_{\text{S}}(\mathbf{z}_t) $.

As shown in Table~\ref{tab:ablation}, the ablation component 'w/ $\mathcal{L}_{\text{S}}(\mathbf{z}_t)$' yields a slightly better motion score; however, it results in a significant decrease in both ID and CPBD scores.
In contrast, 'w/ $ \mathcal{L}_{\text{S}}(\mathbf{z}_0) $' demonstrates superior human structure stability and ID consistency.
We attribute this phenomenon to $\mathcal{L}_{\text{S}}(\mathbf{z}_t)$ destructing more texture information due to the higher noise level.
Consequently, the supervision becomes more biased towards overall human structure movement, capturing better motion patterns.
However, the lack of fine-grained facial texture details makes the supervision effect of $\mathcal{L}_{\text{S}}(\mathbf{z}_t)$ less effective in maintaining ID consistency compared to $\mathcal{L}_{\text{S}}(\mathbf{z}_0)$.
After careful consideration, we select $ \mathcal{L}_{\text{S}}(\mathbf{z}_0) $ as the final implementation to more reliably generate consistent human structures.

\subsubsection{ID Alignment Module Inputs}
Our ID alignment module also has two implementations, differing in whether to utilize depth latents as input.
In case of using depth latents, we concatenate the RGB video latent with the corresponding depth latent along the channel dimension.

To compare the quality of the feature distributions learned by these two implementations, we randomly select $200$ videos and extract corresponding ID representations using each ID alignment module implementation.
We then measure the average intra-video feature distance and the average inter-video feature distance.
The intra-video feature distance reflects the compactness of ID representation distributions within a single video, whereas the inter-video feature distance indicates the distinguishability of ID representation distributions between different characters.
A larger difference between the inter-video and intra-video feature distances demonstrates the effectiveness of the learned ID representations.
The results, shown in Table~\ref{tab:ArcfaceAblation}, indicate that the first method, i.e., utilizing depth latents as input, performs better, as it yields a larger difference between average intra-video and inter-video distances.
We attribute this improvement to the concatenated depth latents, which serve as an implicit facial mask and enhance the module’s ability to detect human faces more effectively.
Consequently, we choose the depth latent concatenation approach as the final implementation for our ID alignment module.

\begin{table}[t]
    \centering
    \begin{tabular}{lcc}
        \hline
        Method & Intra-Video Dist. & Inter-Video Dist. \\
        \hline
        w/ depth input & 0.0417 & 0.1299 \\
        w/o depth input & 0.0487 & 0.1084 \\
        \hline
    \end{tabular}
    \caption{\textbf{Ablation Analysis on ID Alignment Module Implementations.} Intra-Video Dist. represents the average intra-video distance, while Inter-Video Dist. represents the average inter-video distance.}
    \label{tab:ArcfaceAblation}
\end{table}

\subsubsection{Semantic Representation Alignment Supervision}
To demonstrate the effectiveness of our proposed semantic representation alignment supervision, we conducted a detailed ablation analysis, comparing fine-tuning without additional supervision, fine-tuning with $ \mathcal{L}_{\text{S}} $, fine-tuning with $ \mathcal{L}_{\text{ID}} $, and fine-tuning with both $ \mathcal{L}_{\text{ID}}$ and  $\mathcal{L}_{\text{S}}$.
As shown in Table~\ref{tab:ablation}, fine-tuning with $ \mathcal{L}_{\text{S}} $ significantly reduces CPBD and Motion Score values, while fine-tuning with $ \mathcal{L}_{\text{ID}} $ notably improves the ID Score, both without impacting other metrics.
This suggests that our structure representation alignment supervision enables the model to better learn priors on human motion, producing more stable human structures, while ID representation alignment supervision enhances character consistency in generated videos.
Finally, combining both types of supervision, i.e., fine-tuning with both $ \mathcal{L}_{\text{ID}}$ and  $\mathcal{L}_{\text{S}} $, yields the best results in terms of structural stability and ID consistency, validating the effectiveness of our approach.

Furthermore, we investigated the impact of the weights assigned to semantic representation alignment supervision. 
Assigning excessively high weights disrupts the original priors of the video generation model, whereas assigning weights that are too low fails to provide effective supervision.
Therefore, selecting appropriate weights is crucial.
As shown in Table~\ref{tab:ablation}, after thoroughly evaluating the influence on various metrics, we set the weight $\lambda_{\text{S}}$ for the structure loss $\mathcal{L}_{\text{S}}$ to $0.01$ and the weight $\lambda_{\text{ID}}$ for the ID loss $ \mathcal{L}_{\text{ID}}$ to $1$. This configuration achieves a balanced performance across all metrics, attaining state-of-the-art results.
\section{Conclusion}

In this paper, we address the persistent challenges of limb twisting and facial distortion in human image animation, particularly in generating long videos and modeling complex motion.
We introduce a novel approach that leverages semantic representation alignment as supervision rather than as conditional input, preserving generation flexibility while enhancing quality.
Our method incorporates a structure alignment module and an ID alignment module to ensure consistent human structure and identity throughout generated sequences.
By pretraining the structure alignment module on VAE-encoded video latents to predict the structure representations, and using it to supervise the diffusion model, we achieve coherent human structures aligned with ground truth.
The ID alignment module further ensures identity consistency, leveraging predicted structure representations for further enhanced alignment in critical regions.
Quantitative and qualitative evaluations on our curated test set demonstrate the superiority of our approach over current state-of-the-art image-to-video models, with improved performance across multiple metrics, offering a more robust solution for generating long, consistent human motion videos.

{
    \small
    \bibliographystyle{ieeenat_fullname}
    \bibliography{main}
}


\clearpage

\onecolumn
{
    \centering
    \Large
    \textbf{Improving Human Image Animation via Semantic Representation Alignment}\\
    \vspace{0.5em}Supplementary Material \\
}
\setcounter{page}{1}

\section{Implementation Details}
Our base model is CogVideoX 1.0, which uses T5 as the text encoder, with VAE compression ratios of $4$ for temporal and $8 \times 8$ for spatial dimensions.
Our experiments are conducted on 8 NVIDIA A100 GPUs.
We use 8-bit Adam as the optimizer with a learning rate of $1 \times 10^{-5}$.
Both the structure alignment module pretraining and diffusion transformer fine-tuning utilize gradient checkpointing to reduce CUDA memory requirements.
During the pretraining phase of the alignment modules, the batch size is set to $32$ for structure alignment module and $48$ for ID alignment module, while in the diffusion transformer fine-tuning phase, it is set to $8$.
We pretrain the structure alignment module on the collected dataset for $15,000$ steps and the ID alignment module for $2,000$ steps.
Finally, we fine-tune the diffusion transformer with the pretrained alignment modules for $5,000$ steps.
The weights for structure loss and ID loss are set to $0.01$ and $1$, respectively.

\section{Qualitative Visualization}

\subsection{Qualitative Comparison with Other Baselines}

\begin{figure*}[!ht]
  \centering
  \includegraphics[width=\textwidth]{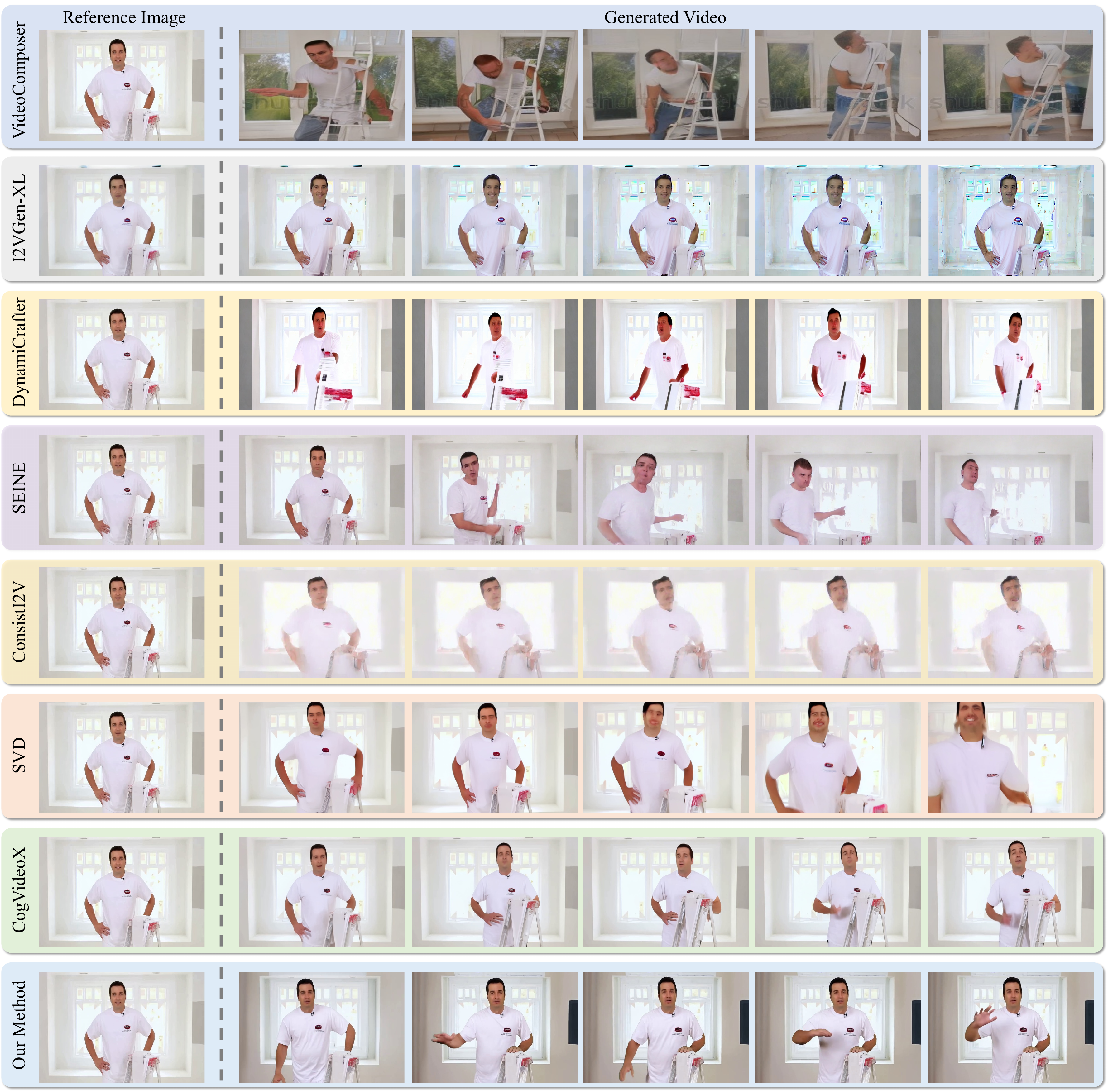} \\
  \vspace{-10pt}
  \caption{
  Qualitative Comparison with other baselines. Our method significantly outperforms other models in terms of human structure stability and ID consistency, effectively avoiding issues of facial distortion and limb twisting. 
    }
    \vspace{-10pt}
 \label{fig:visual}
\end{figure*}

\begin{figure*}[t]
  \centering
  \includegraphics[width=\textwidth]{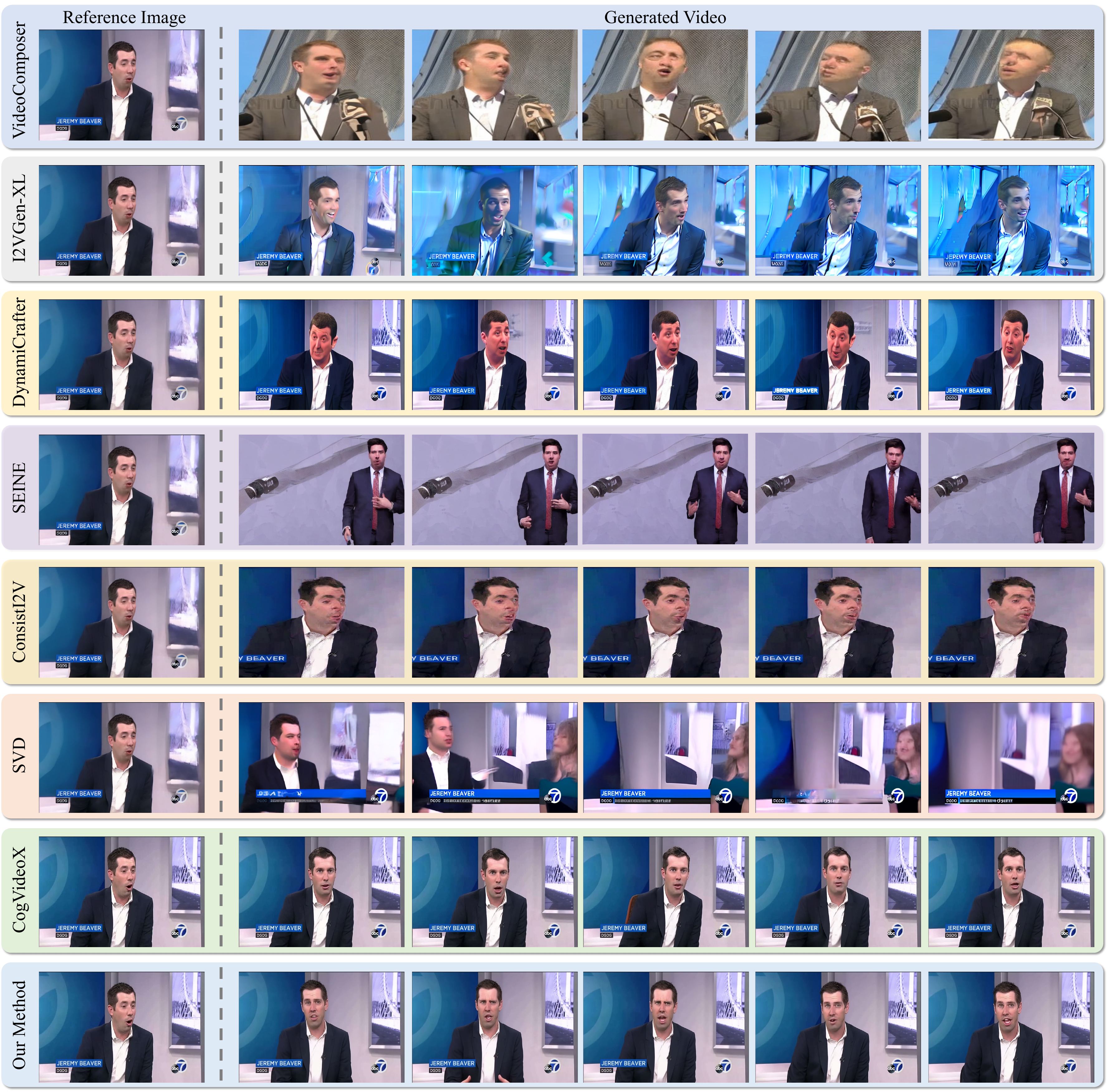} \\
  \caption{
  Qualitative Comparison with Other Baselines. Our method significantly outperforms other models in terms of human structure stability and ID consistency, effectively avoiding issues of facial distortion and limb twisting. 
    }
 \label{fig:visual1}
\end{figure*}

\begin{figure*}[t]
  \centering
  \includegraphics[width=\textwidth]{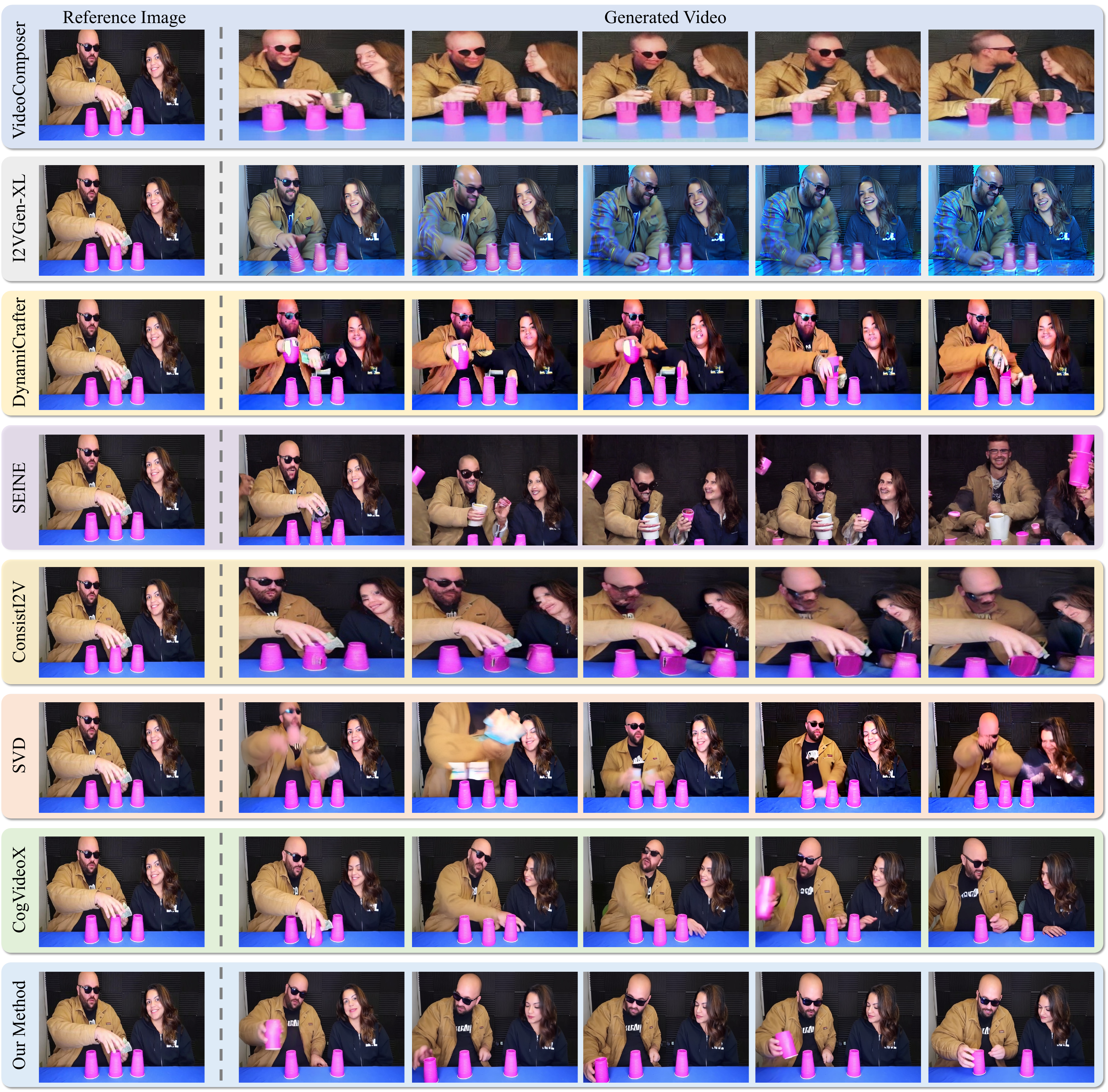} \\
  \caption{
  Qualitative Comparison with other baselines. Our method significantly outperforms other models in terms of human structure stability and ID consistency, effectively avoiding issues of facial distortion and limb twisting. 
    }
 \label{fig:visual2}
\end{figure*}

\begin{figure*}[t]
  \centering
  \includegraphics[width=\textwidth]{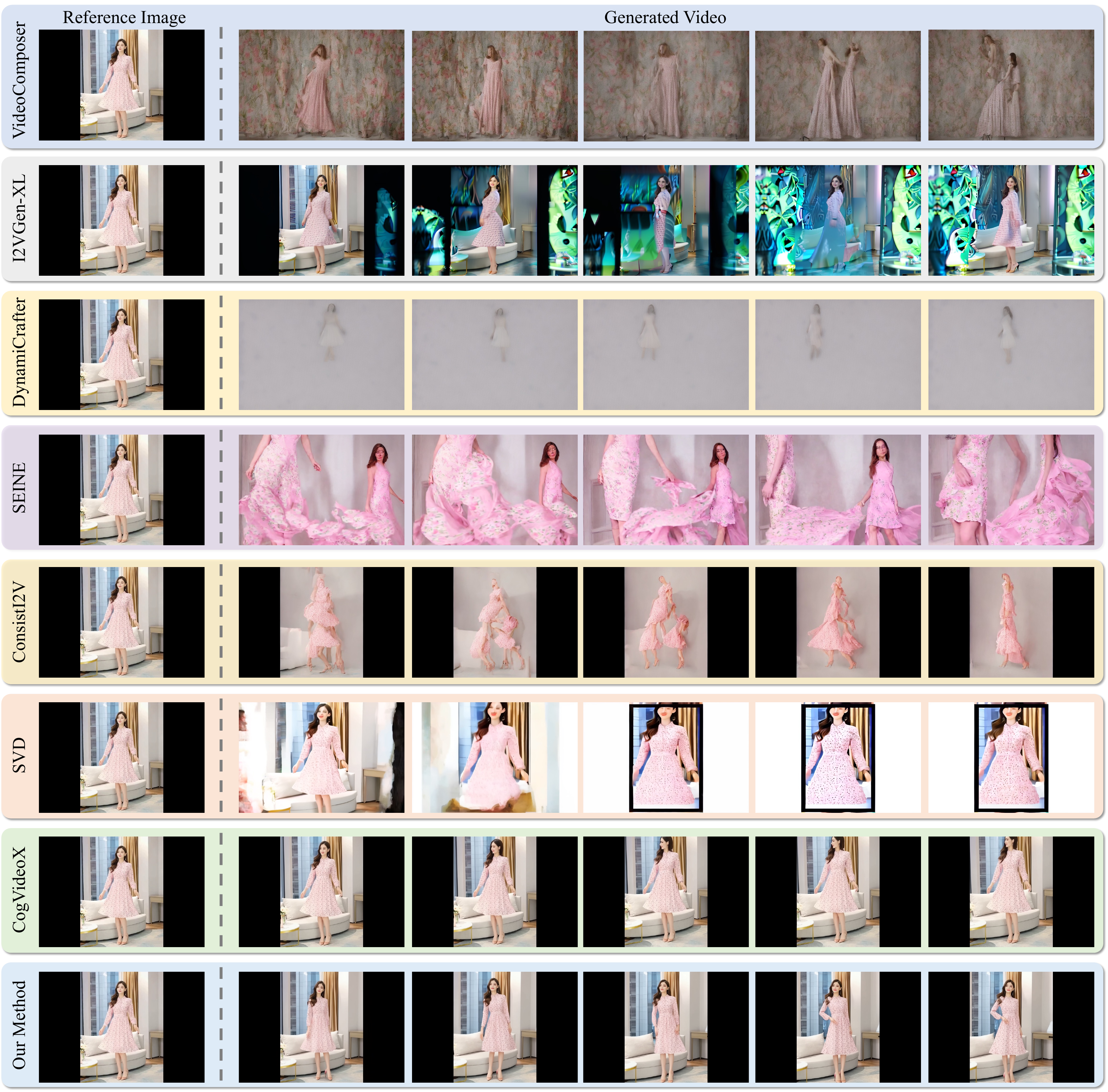} \\
  \caption{
  Qualitative Comparison with other baselines. Our method significantly outperforms other models in terms of human structure stability and ID consistency, effectively avoiding issues of facial distortion and limb twisting. 
    }
 \label{fig:visual3}
\end{figure*}

We conduct qualitative comparison of our method against other baselines.
As illustrated in Figure~\ref{fig:visual}, Figure~\ref{fig:visual1}, Figure~\ref{fig:visual2}, and Figure~\ref{fig:visual3}, our proposed method demonstrate significantly better character consistency and human structure stability.

\clearpage

\subsection{Qualitative Ablation Results}

\begin{figure*}[!h]
  \centering
  \includegraphics[width=\textwidth]{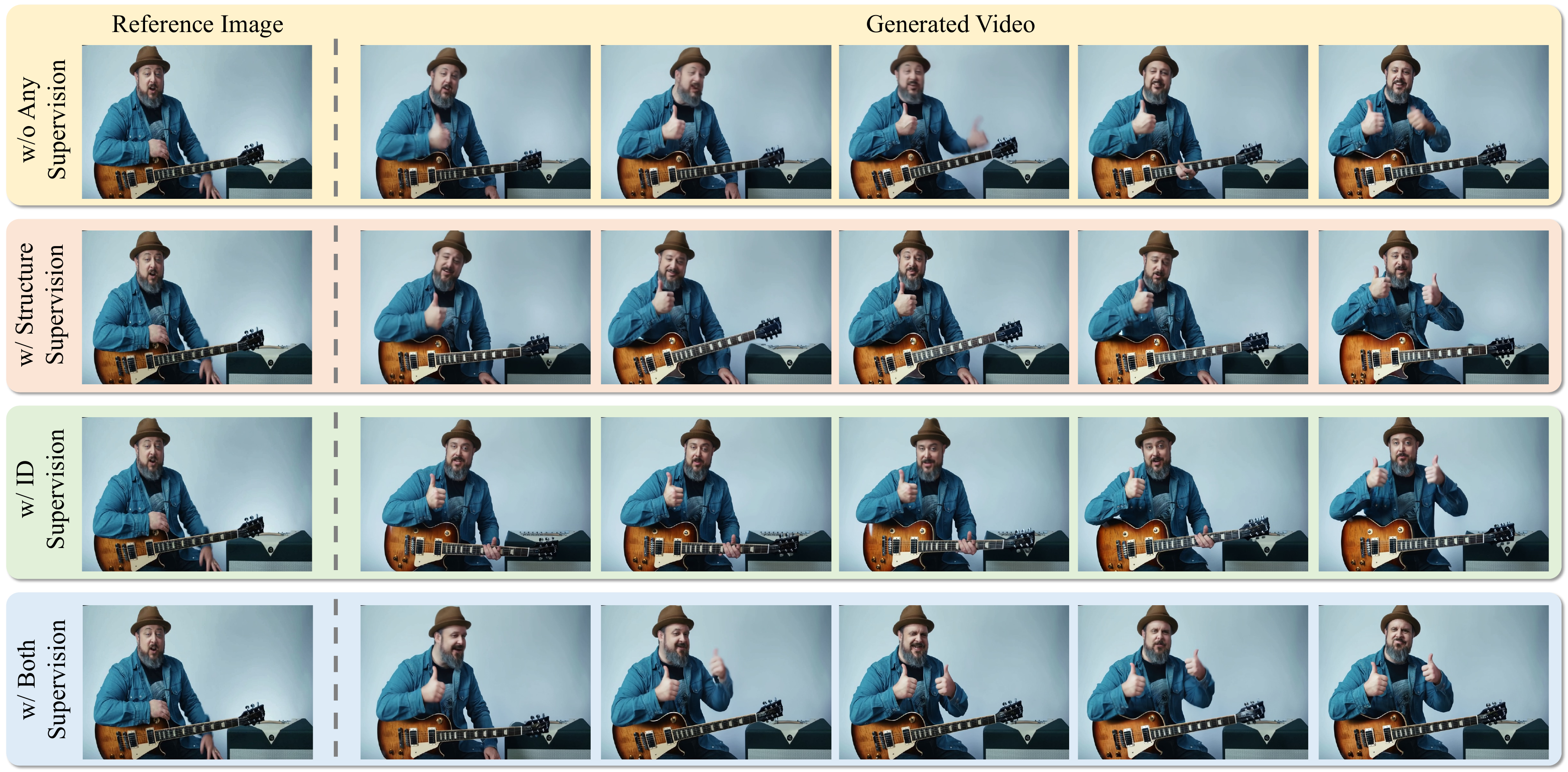} \\
  \caption{
  Qualitative Ablation Comparison Results. Our structure representation alignment supervision enables the model to produce stable human structures, while ID representation alignment supervision enhances character consistency in generated videos. 
    }
 \label{fig:ablation1}
\end{figure*}

\begin{figure*}[!ht]
  \centering
  \includegraphics[width=\textwidth]{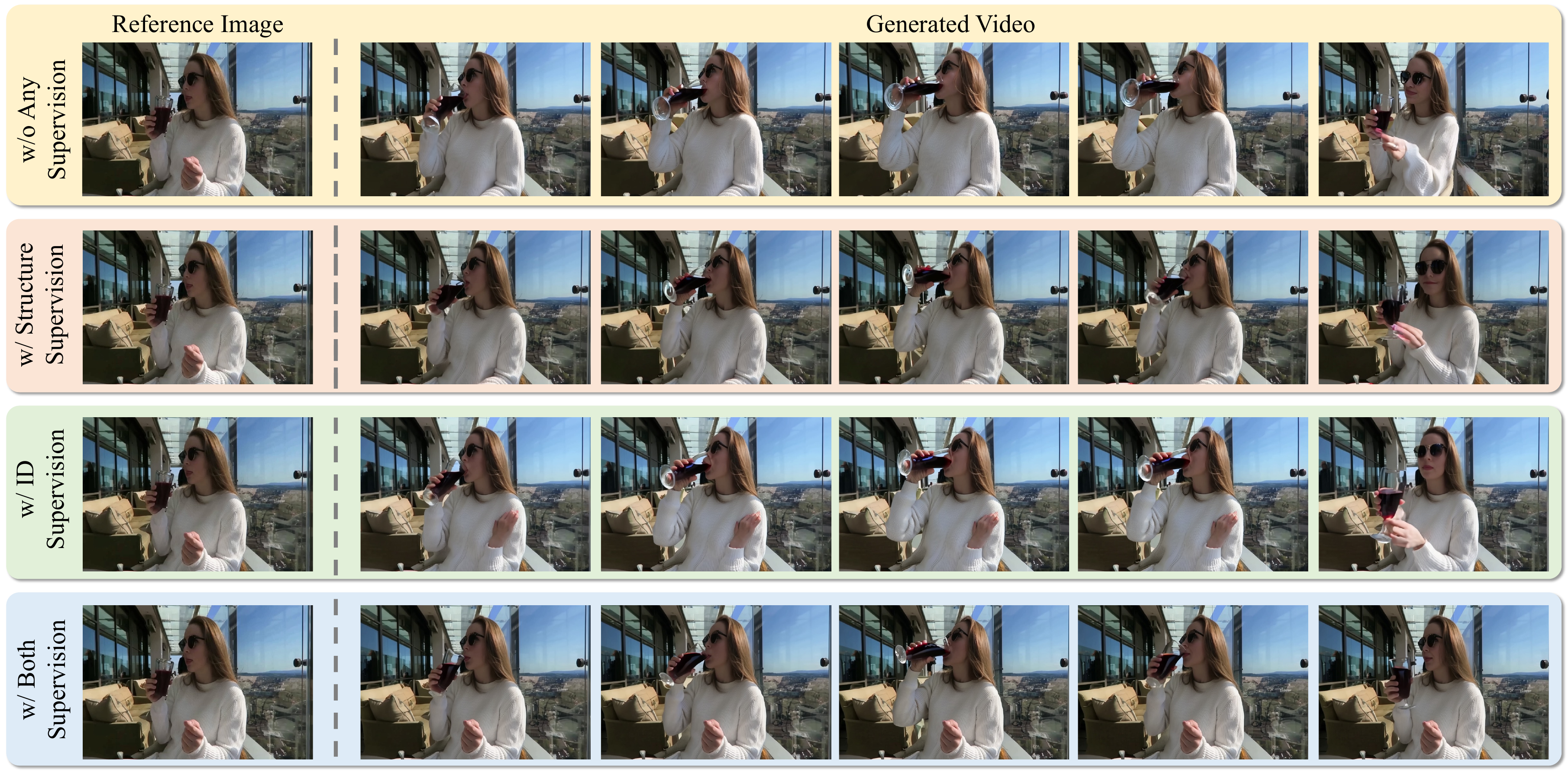} \\
  \caption{
  Qualitative Ablation Comparison Results. Our structure representation alignment supervision enables the model to produce stable human structures, while ID representation alignment supervision enhances character consistency in generated videos. 
    }
 \label{fig:ablation2}
\end{figure*}

\begin{figure*}[!ht]
  \centering
  \includegraphics[width=\textwidth]{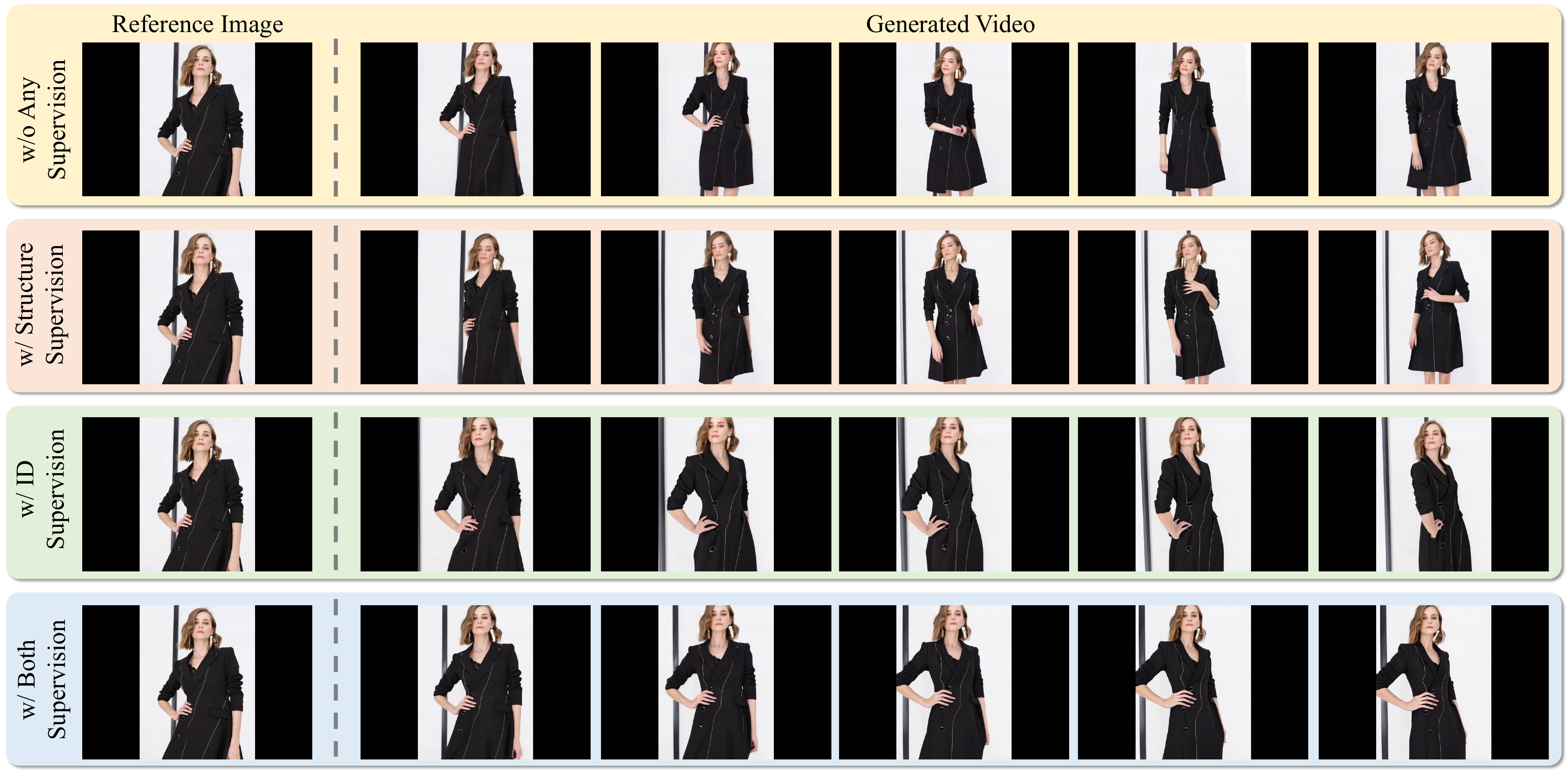} \\
  \caption{
  Qualitative Ablation Comparison Results. Our structure representation alignment supervision enables the model to produce stable human structures, while ID representation alignment supervision enhances character consistency in generated videos. 
    }
 \label{fig:ablation3}
\end{figure*}

As illustrated in Figure~\ref{fig:ablation1}, Figure~\ref{fig:ablation2}, and Figure~\ref{fig:ablation3}, the qualitative ablation analysis suggests that our structure representation alignment supervision allows the model to better capture priors on human motion, resulting in more stable human structures.
Additionally, ID representation alignment supervision enhances character consistency in the generated videos.
Combining both types of supervision, i.e., fine-tuning with both $ \mathcal{L}_{\text{ID}}$ and  $\mathcal{L}_{\text{struc}} $, yields the best results in terms of structural stability and ID consistency, thereby validating the effectiveness of our approach.

\end{document}